\documentclass[runningheads]{llncs}
\usepackage[T1]{fontenc}
\usepackage{graphicx}
\usepackage{multirow} 
\usepackage{algorithm}
\usepackage{algorithmic}
\usepackage{amsmath}
\usepackage{subfig}
\usepackage{float}
\usepackage{hyperref}
\usepackage{booktabs}

\usepackage{fontawesome5}
\usepackage{xcolor}     

\newcommand{\myorcidicon}[1]{\href{https://orcid.org/#1}{\textcolor{green!80!black}{\faOrcid}}}
\usepackage{color}

\begin{document}
\title{Context-Driven Knowledge Graph Completion with Semantic-Aware Relational Message Passing}
\titlerunning{Semantic-Aware Relational Message Passing}

\author{Yan Wen\inst{2}\myorcidicon{0009-0009-4104-269X}\and 
Ruihao Zhou\inst{1}\myorcidicon{0009-0009-9253-9436} \and
Ruitong Liu\inst{3}\myorcidicon{0009-0007-484-7181} \and 
Te Sun\inst{4}\myorcidicon{0009-0006-7840-4955} \and 
Jingyi Kang\inst{1} \and
Yunjia Wu\inst{1}\myorcidicon{0009-0007-1564-4043}\and
Siyuan Li\inst{1}\thanks{Corresponding author.}\myorcidicon{0009-0005-0035-5295}}     
\authorrunning{S. Li et al.}
\institute{School of Computer Science and Technology, Dalian University of Technology, Dalian, Liaoning, China \\
\email{yuanlsy@mail.dlut.edu.cn} \and
Beijing Institute of Technology, Beijing, China\and
School of Mathematical Sciences, Dalian University of Technology, Dalian, Liaoning, China \and
School of Control Science and Engineering, Dalian University of Technology, Dalian, Liaoning, China 
}


%
%
%
\maketitle              
\begin{abstract}
Semantic context surrounding a triplet $(h, r, t)$ is crucial for Knowledge Graph Completion (KGC), providing vital cues for prediction.
However, traditional node-based message passing mechanisms, when applied to knowledge graphs, often introduce noise and suffer from information dilution or over-smoothing by indiscriminately aggregating information from all neighboring edges.
To address this challenge, we propose a semantic-aware relational message passing. A core innovation of this framework is the introduction of a \textbf{semantic-aware Top-K neighbor selection strategy}.
Specifically, this strategy first evaluates the semantic relevance between a central node and its incident edges within a shared latent space, selecting only the Top-K most pertinent ones.
Subsequently, information from these selected edges is effectively fused with the central node's own representation using a \textbf{multi-head attention aggregator} to generate a semantically focused node message.
In this manner, our model not only leverages the structure and features of edges within the knowledge graph but also more accurately captures and propagates the contextual information most relevant to the specific link prediction task, thereby effectively mitigating interference from irrelevant information.
Extensive experiments demonstrate that our method achieves superior performance compared to existing approaches on several established benchmarks.

\end{abstract}
\section{Introduction}

Knowledge graphs (KGs) are structured collections of objective information used to represent real-world entities and their relationships~\cite{li2022kcube,li2022constructing,dong2023hierarchy}. They are typically stored as triplet $(h, r, t)$, where $h$ and $t$ represent the head and tail entities, respectively, and $r$ denotes the relation between them. Real-world knowledge graphs are characterized by their vast size and inherent incompleteness~\cite{wang2017knowledge,ji2021survey}, which poses limitations for their application in downstream tasks. Consequently, automatic knowledge graph completion (KGC) has become a widely investigated research topic within the knowledge graph community~\cite{ji2021survey,chen2020review,zhang2023integrating}.

Embedding-based methods~\cite{bordes2013translating,yang2014embedding,lin2015learning,trouillon2016complex} are dominant knowledge graph completion approaches.
A typical strategy involves embedding entities and relations into low-dimensional vector spaces, each with distinct objectives.
By defining scoring functions within these spaces, these methods learn the underlying link rules between entities and relations and subsequently predict relations between any two given entities.
Although subsequent studies have successfully defined diverse scoring functions in specific vector spaces~\cite{bordes2013translating,sun2019rotate,zhang2020learning} to effectively handle various rule subsets, the intricate nature of real-world KGs results in correspondingly complex rules among entities and relations.
Hence, traditional knowledge graph embedding methods struggle to capture all link rules solely through explicit scoring functions in vector spaces.
At the same time, this explicit embedding of entities and relations leads to the approach being limited to transductive learning and not extending to inductive learning.

Following the success of Graph Neural Networks (GNNs) in modeling graph-structured data, GNNs have been introduced to the structured data within knowledge graphs. Methods like R-GCN, CompGCN, KE-GCN, and PathCon propose updating entity representations by aggregating information from all neighbors in each layer. However, this indiscriminate aggregation can lead to the introduction of excessive noise and is prone to causing the graph over-smoothing problem, where entity representations become indistinguishable after several layers of message passing. Although subsequent methods, such as those conceptually similar to RED-GNN~\cite{zhang2022knowledge}, attempt to mitigate this by selecting strongly relevant edges via soft attention mechanisms, this approach still has inherent limitations. Specifically, as every edge (even with a very low weight) can potentially contribute to the final representation, the risk of noise and information dilution persists, potentially obscuring the truly vital cues offered by the semantic context surrounding a target triplet $(h,r,t)$ for Knowledge Graph Completion.

To address these aforementioned challenges more effectively and to better leverage the crucial semantic context surrounding a target triplet $(h,r,t)$, we propose a semantic-aware relational message passing, significantly enhanced by a \textbf{semantic-aware Top-K neighbor selection strategy}. Instead of assigning continuous weights to all incident edges, our strategy first evaluates the semantic relevance between a node and its incident edges within a shared latent space, and then selects only the Top-K most pertinent ones. Subsequently, information from these carefully selected Top-K edges is then effectively fused with the node's own representation using a \textbf{multi-head attention aggregator} to generate a semantically focused node message.

Our contributions can be summarized as follows:
\begin{itemize}
    \item We propose a \textbf{semantic-aware relational message passing} that enhances completion performance by focusing on \textbf{semantic relevance} in message construction.
    \item We introduce a novel \textbf{Top-K semantic-aware neighbor selection strategy} that prioritizes semantically relevant edges based on a learnable similarity score, thereby reducing noise and information dilution.
    \item We develop a \textbf{specialized multi-head attention aggregator} to effectively fuse a node's representation with information from its semantically selected neighborhood, generating more context-rich node messages.
    \item We demonstrate through comprehensive experiments that our approach achieves state-of-the-art performance across all datasets, validating its robustness in link prediction across diverse knowledge graphs.
\end{itemize}

\section{Related Works}

\subsection{Knowledge Graph Completion}
Knowledge Graphs (KGs) structurally store real-world information as triplets $(h, r, t)$~\cite{li2022kcube,li2022constructing,dong2023hierarchy}, but their inherent incompleteness limits their application~\cite{wang2017knowledge,ji2021survey}.
Knowledge Graph Completion aims to address this issue~\cite{ji2021survey,chen2020review,zhang2023integrating}.
Embedding-based methods are dominant in KGC~\cite{bordes2013translating,yang2014embedding,lin2015learning,trouillon2016complex}, mapping entities and relations to low-dimensional vector spaces and learning link rules via scoring functions.
Although subsequent studies have proposed diverse scoring functions~\cite{bordes2013translating,sun2019rotate,zhang2020learning}, these methods still struggle to capture all complex rules through explicit scoring functions and are limited to transductive learning.

\subsection{Graph Neural Networks}
Graph Neural Networks (GNNs)~\cite{zhang2022knowledge,dettmers2018convolutional,vashishth2020composition,schlichtkrull2018modeling,zhang2024logical} have been introduced to KGs due to their strong modeling capabilities on graph-structured data.
Methods like R-GCN~\cite{schlichtkrull2018modeling} and CompGCN~\cite{vashishth2020composition} update entity representations by aggregating information from neighbors.
However, this indiscriminate aggregation can introduce excessive noise and may lead to the over-smoothing problem, where entity representations become indistinguishable after multiple message-passing layers.
While some subsequent methods (e.g., conceptually similar to RED-GNN~\cite{zhang2022knowledge}) attempt to mitigate this using soft attention mechanisms to select edges, the risk of noise and information dilution persists as all edges (even with low weights) can still contribute to the final representation, potentially obscuring crucial semantic cues around a target triplet $(h,r,t)$.

\section{Methodology}

\subsection{Preliminaries}
A knowledge graph is represented as \( K = (\mathcal{V}, \mathcal{E}, \mathcal{F}) \), where \( \mathcal{V} \) is the set of entities, \( \mathcal{E} \) is the set of relations, and \( \mathcal{F} = \{(h, r, t) \mid h, t \in \mathcal{V}, r \in \mathcal{E}\} \) is the set of factual triplets.
Given a query entity \( h \), a query relation \( r \), and an unknown answer entity \( t \), the task of knowledge graph completion is to predict \( t \) for the query \((h, r, ?)\).
Typically, any entity in \( \mathcal{V} \) is a candidate for \( t \).

Our objective is to model the probability distribution of the tail entity, $p(t|h, r)$.
Applying Bayes' theorem, this can be expressed as:
\begin{equation}
    p(t|h, r) \propto p(h, r|t) \cdot p(t), \label{eq:bayes_tail_entity_v3}
\end{equation}
where $p(t)$ is the prior distribution of the tail entity.
The term $p(h, r|t)$ can be further decomposed into a symmetric form:
\begin{equation}
    p(h, r|t) = \frac{1}{2} \left[ p(h|t) \cdot p(r|h, t) +
    p(r|t) \cdot p(h|r, t) \right], \label{eq:decomposition_tail_entity_v3}
\end{equation}
This decomposition guides our model design.
The terms $p(h|t)$ and $p(r|t)$ assess the likelihood of head entity $h$ or relation $r$ given the tail entity $t$. Our approach models these by emphasizing the semantic relevance of the contextual information. When considering the neighborhood information surrounding $h$ and $r$, our Top-K semantic-aware neighbor selection mechanism is employed to identify and prioritize the most pertinent contextual elements with respect to $t$.
The terms $p(r|h, t)$ and $p(h|r, t)$ measure the likelihood of $h$ being connected to $t$ via $r$. These are modeled by learning effective entity and relation representations that capture potential connection patterns. Our semantic-aware relational message passing, enhanced by the Top-K selection and a specialized multi-head attention aggregator, is designed to generate context-rich representations by effectively fusing information from these semantically selected neighborhoods.
Below, we will detail how these components synergistically model these probabilistic factors to achieve link prediction.

\subsection{semantic-aware relational message passing} 

The semantic context surrounding a triplet $(h, r, t)$ provides crucial clues for knowledge graph completion. However, traditional node-based message passing, when applied to knowledge graphs, may introduce noise and lead to information dilution or over-smoothing due to the indiscriminate aggregation of all neighboring nodes. Meanwhile, we observe that edges in knowledge graphs also possess corresponding features and exhibit certain inferential properties. For instance, given $(A, \textit{is father of}, B) \land (C, \textit{is wife of}, A) \rightarrow (C, \textit{is mother of}, B)$, the inference here does not focus on node attributes but relies more heavily on relational features. More importantly, the number of relations in the real world is significantly smaller than the number of entities. Therefore, relation-based message passing is more efficient and can provide richer and more guiding clues.

\begin{figure}
    \centering
\includegraphics[width=1\linewidth,
                     alt={The architecture of the semantic-aware relational message passing model. The left side illustrates the overall workflow: a central edge selects its Top-K neighbors based on semantic similarity, aggregates their information, and updates its state. The right side shows a detailed diagram of the multi-head attention aggregator, which processes the central edge's state and the aggregated neighbor state to produce the final output.}]
                    {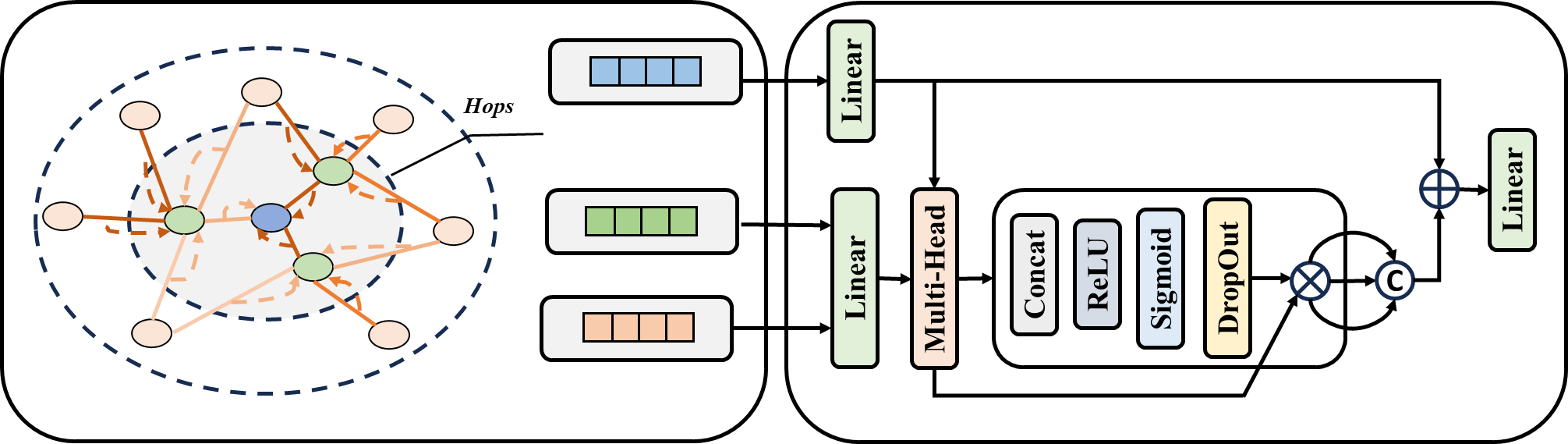}
    \caption{A framework for semantic-aware relational message passing, with message passing and aggregation represented on the left, and the structure of a multi-head attention aggregator on the right}
    \label{fig:enter-label}
\end{figure}

To this end, we adopt a semantic-aware relational message passingc, as shown in figure~\ref{fig:enter-label}, with its core concept inspired by PathCon~\cite{wang2021relational}. We enhance its knowledge representation and reasoning capabilities by introducing a semantic-aware Top-K neighbor selection strategy and a carefully designed multi-head attention aggregator. In the following, we will elaborate on these components in detail.

A cornerstone of our enhanced message passing paradigm is the \textbf{semantic-aware Top-K edge selection strategy}, which refines how a node's representation is updated. The entire process begins by contextually enriching each edge $e_v \in \mathcal{N}(v)$ incident to a target node $v$. For each such edge, treated as a central edge, we update its state by selectively aggregating information from its own local neighborhood of edges, $\mathcal{N}(e_v)$. This mitigates noise and information dilution inherent in indiscriminate aggregation. Specifically, for a central edge $e_v$ with its current state $\mathbf{s}_{e_v}^{(l)}$, we assess its semantic relevance to each neighboring edge $e_n \in \mathcal{N}(e_v)$ by projecting their states into a shared latent semantic space via a learnable mapping function $f(\cdot)$. Within this space, their similarity score is computed as:
\begin{equation}
    \text{Score}(e_v, e_n) = \exp(-\|f(\mathbf{s}_{e_v}^{(l)}) - f(\mathbf{s}_{e_n}^{(l)})\|^2 / \tau),
    \label{eq:semantic_score_re}
\end{equation}
where $\| \cdot \|^2$ denotes the squared Euclidean distance, and $\tau$ is a temperature hyperparameter. Only the Top-K edges $\mathcal{N}_K(e_v) \subseteq \mathcal{N}(e_v)$ exhibiting the highest scores are chosen. The states of these selected edges, $\{\mathbf{s}_{e_n}^{(l)}\}_{e_n \in \mathcal{N}_K(e_v)}$, are first consolidated into a unified neighborhood representation through mean aggregation:
\begin{equation}
    \mathbf{\bar{s}}_{\mathcal{N}_K(e_v)}^{(l)} = \text{Mean}\left( \{\mathbf{s}_{e_n}^{(l)}\}_{e_n \in \mathcal{N}_K(e_v)} \right).
    \label{eq:mean_agg_edges_re}
\end{equation}
This aggregated representation and the central edge's own state are then processed by a multi-head attention aggregator to produce a contextually-aware, updated edge state $\mathbf{s}_{e_v}^{(l+1)}$. After this enrichment procedure is performed for every edge incident to node $v$, we obtain a set of updated edge representations $\{\mathbf{s}_{e_v}^{(l+1)}\}_{e_v \in \mathcal{N}(v)}$. The final updated representation for the node, $\mathbf{h}_v^{(l+1)}$, is then derived by applying mean pooling to these enriched incident edge states. This approach defines the node's identity purely through the contextualized information of its surrounding relational structure, without direct use of its previous state $\mathbf{h}_v^{(l)}$:
\begin{equation}
    \mathbf{h}_v^{(l+1)} = \text{Mean}\left(\{\mathbf{s}_{e_v}^{(l+1)}\}_{e_v \in \mathcal{N}(v)}\right).
    \label{eq:final_node_update_re}
\end{equation}
The impact of the semantic scoring function and the aggregator selection will be further discussed in subsequent ablation studies.

\subsubsection{Alternating Relational Message Passing.}
 This scheme iteratively refines both node and edge representations. Following the node representation update detailed above, messages for edges are constructed. For an edge $e=(u,v)$, its message $\mathbf{m}_e^{(l+1)}$ is derived from the newly updated representations of its endpoint nodes:
\begin{equation}
    \mathbf{m}_e^{(l+1)} = \text{MLP}([\mathbf{h}_u^{(l+1)}, \mathbf{h}_v^{(l+1)}]),
    \label{eq:edge_msg_from_nodes_re}
\end{equation}
where $[\cdot]$ denotes concatenation. This edge message $\mathbf{m}_e^{(l+1)}$ is subsequently used to update the hidden state of edge $e$:
\begin{equation}
    \mathbf{s}_e^{(l+1)} = \sigma \left( \text{Linear}\left(\left[ \mathbf{s}_e^{(l)}, \mathbf{m}_e^{(l+1)}\right]\right) \right),
    \label{eq:edge_state_update_re}
\end{equation}
where $\sigma$ is an activation function. The initial hidden state of an edge is its feature embedding, $\mathbf{s}_e^{(0)} = \mathbf{x}_e$. This alternating process facilitates the propagation of refined, semantically relevant information throughout the graph structure.

In this manner, our model, by integrating the semantic-aware Top-K edge selection and the multi-head attention aggregator within an alternating relational message passing framework, not only leverages the structure and features of edges in the knowledge graph but also more accurately captures and propagates context information most relevant to the specific link prediction task. 

\subsection{Training And Inference}

First, we compute the final messages $\mathbf{m}_h$ and $\mathbf{m}_t$ for the head entity $h$ and tail entity $t$ using a semantic-aware relational message passing. These messages encapsulate their respective semantic contexts.
Subsequently, the score for a triplet $(h, r, t)$ is defined as:
\begin{equation}
    s_{(h,r,t)} = \sigma \left( \left[ \mathbf{m}_h, \mathbf{x}_e, \mathbf{m}_t \right]^T \mathbf{W} + \mathbf{b} \right).
    \label{eq:context_representation}
\end{equation}
Here, $[\mathbf{m}_h, \mathbf{x}_e, \mathbf{m}_t]$ denotes the concatenation of the head entity message, the relation representation, and the tail entity message.

Given the effectiveness of alternating positive and negative sampling in learning knowledge graph embeddings and word embeddings \cite{mikolov2013distributed}, we optimize the model using the following negative log-likelihood loss:
\begin{equation}
L = -\log \sigma \left( s_{(h,r,t)} - \gamma \right) - \sum_{i=1}^{n} \frac{1}{n} \log \sigma \left( \gamma - s_{(h,r,t'_i)} \right),
    \label{eq:loss_function}
\end{equation}

where $\gamma$ is a fixed margin, and $\sigma$ denotes the sigmoid function.
In this formulation, $(h, r, t'_i)$ represents the $i$-th negative triplet, generated by corrupting the tail entity of the positive triplet $(h, r, t)$.
A similar formulation is used for head entity prediction, where negative samples $(h'_i, r, t)$ are generated by corrupting the head entity.

The inference process is analogous to training, involving the evaluation of a given triplet's plausibility using Equation~\eqref{eq:context_representation}.

\section{Experiment}

To evaluate the effectiveness of our Semantic-Aware Relational Message Passing(SARMP), we have designed a series of experiments to address the following research questions:
\begin{itemize}
    \item \textbf{RQ1:} How does the performance of SARMP compare to a diverse range of mainstream completion models?
    \item \textbf{RQ2:} What distinct contributions do the key components of SARMP offer to the overall performance? Additionally, how does the model's performance adapt and respond to variations in hyperparameter settings?
    \item \textbf{RQ3:} How does SARMP perform in terms of its number of parameters?
\end{itemize}

\subsection{Experiment Settings}
\paragraph{Datasets.}
We conduct experiments on four standard knowledge graph datasets: (i) FB15K-237, a subset of FB15K with inverse relations removed; (ii) WN18RR, a subset of WN18 with inverse relations removed; (iii) Kinship, a knowledge graph describing kinship relations, focusing on family and blood relations; and (iv) UMLS, a biomedical knowledge graph encompassing medical terminology, concepts, and their interrelationships.
\paragraph{Baselines.}
 Based on the concept of Transductive Knowledge Graph Completion, we categorize all baseline methods into three groups: \textbf{(i) Embedding-based:} TransE~\cite{bordes2013translating}, DistMult~\cite{yang2014embedding}, ComplEx~\cite{trouillon2016complex}, RotatE~\cite{sun2019rotate}, ConvE~\cite{dettmers2018convolutional}, TDN~\cite{wang2023tdn}, and FDM~\cite{long2024fact}; \textbf{(ii) Rule-based:} RNNlogic~\cite{qu2020rnnlogic} and DRUM~\cite{sadeghian2019drum}; \textbf{(iii) GNN-Based:} CompGCN~\cite{vashishth2019composition}, PathCon~\cite{wang2021relational}, RED-GNN ~\cite{zhang2022knowledge} and NBFNet~\cite{zhu2021neural}.

\paragraph{Evaluation Metrics.}
For each test triplet $(h, r, t)$, we construct two queries: $(h, r, ?)$ and $(?, r, t)$, where the answers are $t$ and $h$, respectively. Mean Reciprocal Rank (MRR) and Hits@N are selected as evaluation metrics under the filtered setting~\cite{sun2019rotate}, with definitions consistent with prior research. Higher MRR and H@N scores signify superior model efficacy.

\subsection{Overall Performance Comparison: RQ1}

\begin{table*}
  \centering
    \caption{Performance comparison on FB15k-237, WN18RR, Kinship and UMLS datasets. The best results are shown in bold while the second-best results are shown in the underline.}
  \setlength{\tabcolsep}{2.8pt}
  \renewcommand{\arraystretch}{0.9}
  \footnotesize
\label{tab:Q1_Result} 
  \begin{tabular}{c|c|cc|cc|cc|cc}
    \toprule
    \multirow{2}{*}{\textbf{Type}} & \multirow{2}{*}{\textbf{Model}} & \multicolumn{2}{c|}{\textbf{FB15k-237}} & \multicolumn{2}{c|}{\textbf{WN18RR}} & \multicolumn{2}{c|}{\textbf{Kinship}}  & \multicolumn{2}{c}{\textbf{UMLS}}\\
    \cmidrule(lr){3-4} \cmidrule(lr){5-6} \cmidrule(lr){7-8} \cmidrule(lr){9-10}
                                & & \textbf{MRR} & \textbf{H@1} & \textbf{MRR} & \textbf{H@1} & \textbf{MRR} & \textbf{H@1} & \textbf{MRR} & \textbf{H@1} \\
    \midrule
    \multicolumn{1}{c|}{\multirow{7}{*}{\centering \textbf{Emb-based}}} & TransE                 & 0.289 & 19.8 & 0.226 & -    & 0.310 & 0.9  & 0.690 & 52.3 \\
     & DisMult                   & 0.241 & 15.5 & 0.430 & 39.0 & 0.354 & 18.9 & 0.391 & 25.6 \\
      & ComplEx                   & 0.247 & 15.8 & 0.440 & 41.0 & 0.418 & 24.2 & 0.411 & 27.3 \\
        & RotaE                   & 0.337 & 24.1 & 0.447 & 42.8 & 0.651 & 50.4 & 0.744 & 63.6 \\
   & ConvE                  & 0.325 & 23.7 & 0.430 & 40.0 & 0.685 & 55.2 & 0.756 & 69.7 \\
  & TDN                  & 0.350 & 26.3 & 0.481 & 43.9 & 0.780 & 67.7 & 0.860 & 82.4 \\
    &FDM                   & \underline{0.485} & \underline{38.6} & 0.506 & 45.6 & \underline{0.837} & \underline{76.1} & 0.922 & 89.3 \\
    \midrule
    \multicolumn{1}{c|}{\multirow{2}{*}{\centering \textbf{Rule-based}}}
    & RNNLogic                 & 0.344 & 25.2 & 0.483 & 44.6 & 0.722 & 59.8 & 0.842 & 77.2 \\
   & DRUM                   & 0.238 & 17.4 & 0.382 & 36.9 & 0.334 & 18.3 & 0.548 & 35.8 \\
     \midrule
    \multicolumn{1}{c|}{\multirow{4}{*}{\centering \textbf{GNN-based}}}
    & CompGCN                  & 0.355 & 26.4 & 0.479 & 44.3 & -     & -    & 0.927 & 86.7 \\
   & PathCon                  & 0.483 & 42.5 & 0.522 & 46.2 & -     & -    & -     & -    \\
    & RED-GNN                  & 0.374 & 28.3 & 0.533 & 48.5 & -     & -    & \textbf{0.964} & \textbf{94.6} \\
   & NBFNet                   & 0.415 & 32.1 & \textbf{0.551} & \textbf{49.7} & 0.606 & 43.5 & 0.778 & 68.8 \\
    \midrule
    \multicolumn{1}{c|}{\centering \textbf{Ours}} & \textbf{SARMP}  & \textbf{0.492} & \textbf{44.0} & \underline{0.535} & \underline{49.3} & \textbf{0.863} & \textbf{77.3} & \underline{0.944} & \underline{90.9} \\
    \bottomrule
  \end{tabular}
\end{table*}

\begin{table}[ht]
    \centering
        \caption{Entity prediction by relation category on FB15k-237.}
        \label{tab2}
    \begin{tabular}{lcccccc}
    \toprule
    &  \textbf{TransE} &  \textbf{RotatE} & \textbf{CompGCN} &  \textbf{NBFNet} & \textbf{FDM} & \textbf{SARMP}\\
    \midrule
        \textbf{Head Pred} & & & & & & \\ 
        \multicolumn{1}{c}{1-1}  & 0.498 & 0.487 & 0.457 & \textbf{0.578} & 0.569 & \underline{0.573}  \\
        \multicolumn{1}{c}{1-N}  & 0.079 & 0.081 & 0.112 & 0.165 & \underline{0.203} & \textbf{0.219}  \\
        \multicolumn{1}{c}{N-1}  & 0.455 & 0.467 & 0.471 & 0.499 & \underline{0.559} & \textbf{0.571}  \\
        \multicolumn{1}{c}{N-N}  & 0.224 & 0.234 & 0.275 & 0.348 & \underline{0.423} & \textbf{0.435} \\
    \midrule
    \textbf{Tail Pred} & & & & & & \\
        \multicolumn{1}{c}{1-1}  & 0.488 & 0.484 & 0.453 & \textbf{0.600} & 0.519 & \underline{0.522}  \\
        \multicolumn{1}{c}{1-N}  & 0.744 & 0.747 & 0.779 & 0.790 & \underline{0.826} & \textbf{0.847}  \\
        \multicolumn{1}{c}{N-1}  & 0.071 & 0.070 & 0.076 & 0.122 & \underline{0.167} & \textbf{0.183}  \\
        \multicolumn{1}{c}{N-N}  & 0.330 & 0.338 & 0.395 & 0.456 & \underline{0.543} & \textbf{0.551}  \\
    \bottomrule
    \end{tabular}
\end{table}

To answer RQ1, we perform extensive experiments to compare semantic-aware relational message passing with the baseline models to the best of our knowledge. The results are summarized in Table~\ref{tab:Q1_Result}. Our framework shows remarkable improvement across all metrics on all four datasets. Specifically, it achieves the best MRR scores on the FB15k-237 and Kinship datasets, with improvements of 0.7\% (1.4\% relative) and 2.6\% (3.1\% relative), respectively. On the WN18RR and UMLS datasets, it achieves the second-best MRR and Hit@1 scores.

To further analyze the performance of SARMP, we break it down by the categories of relations~\cite{wang2014knowledge} defined in FB15k-237. Table~\ref{tab2} presents the prediction MRR scores for each category. The results indicate that SARMP achieves a greater relative improvement in the 1-N, N-1 and N-N types. 

In summary, the results demonstrate that our model accurately captures and propagates the most relevant semantic contextual information for the prediction of triples $(h, r, t)$. This context-driven knowledge graph completion paradigm undoubtedly showcases its powerful capabilities.

\subsection{Ablation Study: RQ2}
\subsubsection{Key Modules Ablation Study.}
To verify the effectiveness of our proposed modules, we develop two distinct model variants:
\begin{itemize}
    \item \textbf{w/o Top-K}: We replace the semantic-aware Top-K neighbor selection strategy with a simple random sampling strategy.
    \item \textbf{w/o Score}: We removed the mapping-based similarity scoring, and instead use dot product similarity for scoring. 
\end{itemize}

\begin{table}
  \centering
  \setlength{\tabcolsep}{10pt}
  \caption{Ablation study on key components of SARMP.}
\label{tab:ablation_SARMP}
  \begin{tabular}{l|cc|cc}
    \toprule
    \multicolumn{1}{c|}{\raisebox{-\totalheight}{\textbf{Ablation Settings}}} & \multicolumn{2}{c|}{\textbf{FB15k-237}} & \multicolumn{2}{c}{\textbf{WN18RR}} \\
    \cmidrule(lr){2-3} \cmidrule(lr){4-5}
                               & \textbf{MRR} & \textbf{Hit@1} & \textbf{MRR} & \textbf{Hit@1} \\
    \midrule
    SARMP                         &\textbf{0.492}  &\textbf{44.0}  &\textbf{0.535}  &\textbf{49.3}  \\
    SARMP w/o Top-K               &0.473  &41.8  &0.509  &45.4  \\
    SARMP w/o Score               &0.482  &42.9  &0.521  &46.0  \\
    \bottomrule
  \end{tabular}
\end{table}

The ablation experiments, detailed in Table \ref{tab:ablation_SARMP}, yield several critical insights.
Primarily, the semantic-aware Top-K neighbor selection mechanism has the most pronounced effect on overall model effectiveness; its exclusion leads to a considerable drop in performance, highlighting its essential function in isolating relevant information and curtailing noise.
Additionally, the mapping-based similarity scoring function is also identified as vital for assessing and prioritizing candidate answers, as its removal incurs a clear decrease in performance.
Thus, the collaborative operation of these elements is a cornerstone of the SARMP model's capacity to sustain its state-of-the-art results.

\begin{figure*}[t] 
    \centering
    \begin{minipage}{0.33\linewidth}
        \centering
        \includegraphics[width=\linewidth,
                         alt={A line chart showing model performance (MRR and Hits@1) for different Top-K values. Performance peaks at K equals 10 and then declines.}]
                        {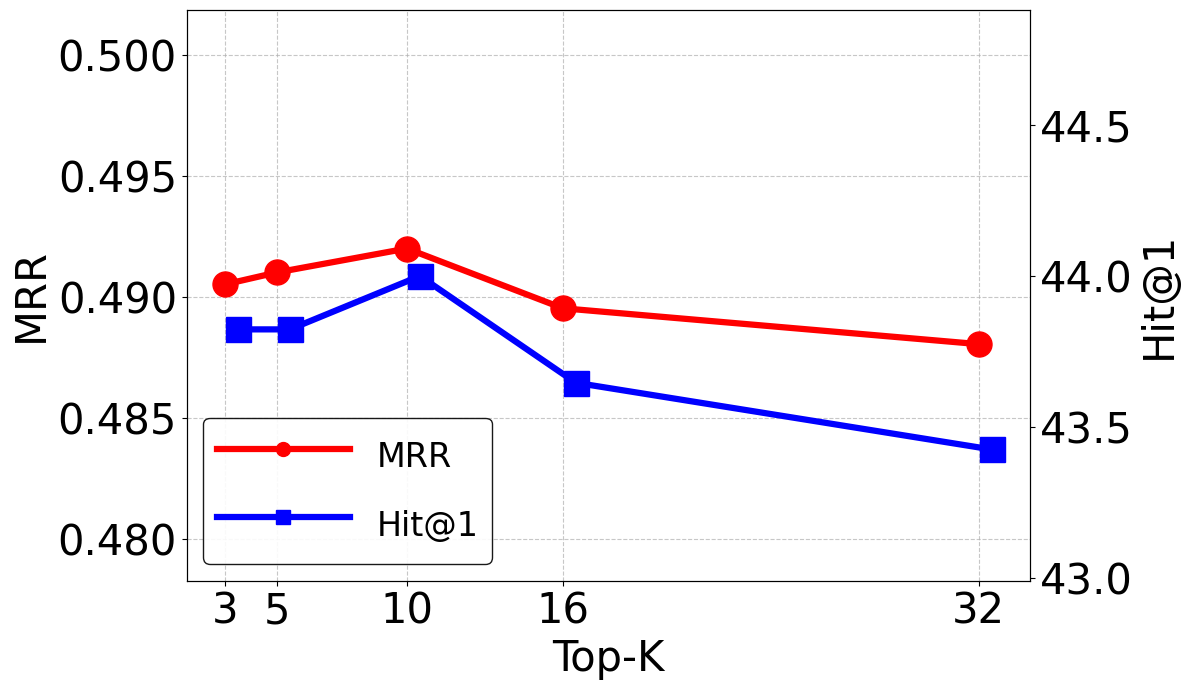}
        \caption*{(a) Results with different Top-K.}
    \end{minipage}\hfill
    \begin{minipage}{0.33\linewidth}
        \centering
        \includegraphics[width=\linewidth,
                         alt={A bar chart comparing performance for different context aggregators. The Multi-Head Attention aggregator achieves the highest scores for both MRR and Hits@1.}]
                        {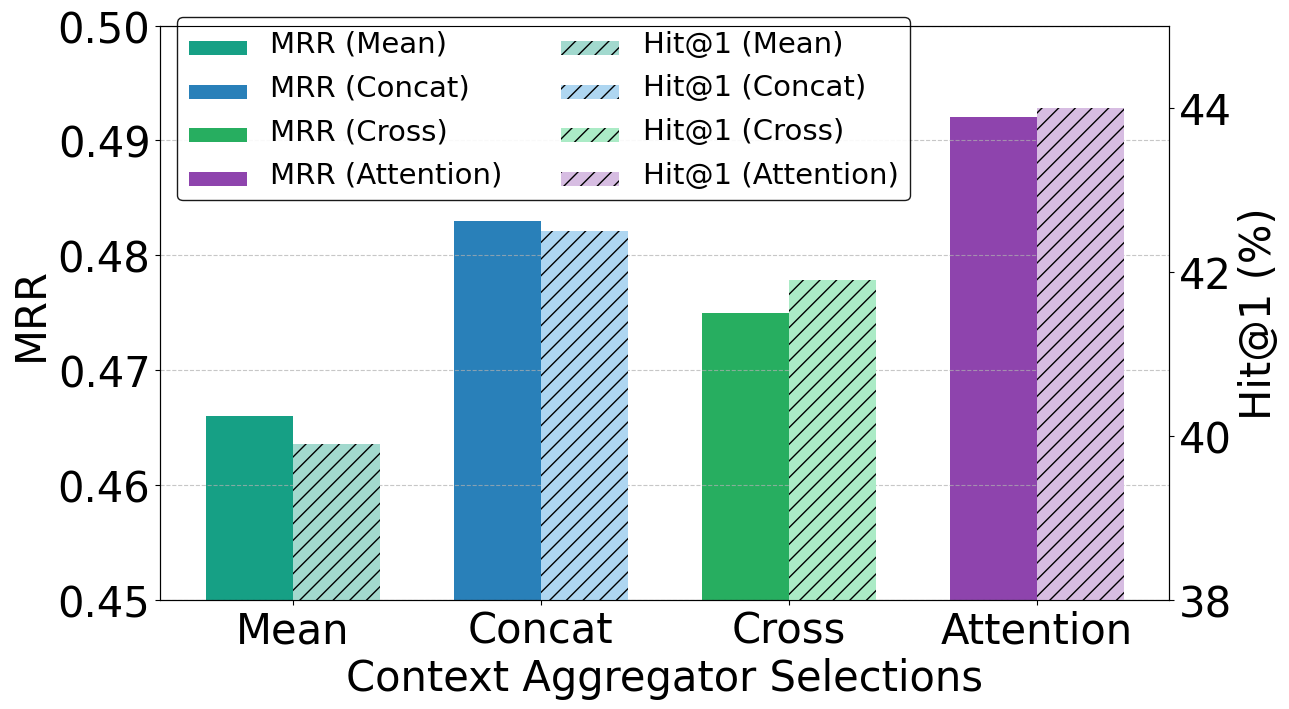}
        \caption*{(b) Results with different context aggregators.}
    \end{minipage}\hfill
    \begin{minipage}{0.33\linewidth}
        \centering
        \includegraphics[width=\linewidth,
                         alt={A line chart showing model performance versus the number of hops. The performance is optimal at 2 hops and degrades with more hops.}]
                        {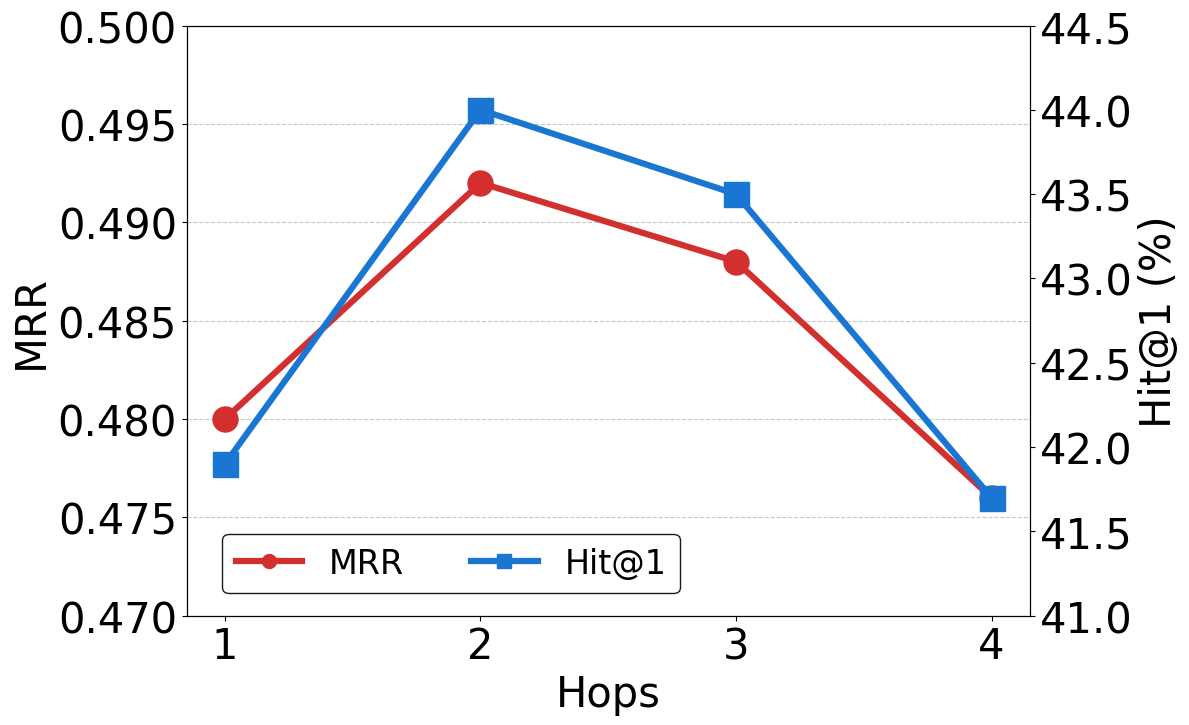}
        \caption*{(c) Results with different hops.}
    \end{minipage}
    \caption{Hyper-parameter analysis of SARMP on FB15k-237.}
    \label{fig:hyperparameter_analysis}
\end{figure*}
\subsubsection{Sensitivity to Key Hyperparameters.}
This study examines the sensitivity of SARMP to key hyperparameters, specifically the Top-K value for semantic neighbor selection, the choice of context aggregator, and the number of hops for graph traversal. Results on the FB15k-237 dataset are presented in \autoref{fig:hyperparameter_analysis}.

\autoref{fig:hyperparameter_analysis}(a) illustrates SARMP's performance across different Top-K values. A moderate K (specifically 10 in this case) is crucial. A small K provides insufficient contextual information, while a large K can introduce noise from excessive, less relevant information, thereby reducing performance.

\autoref{fig:hyperparameter_analysis}(b) compares SARMP's performance with various context aggregators. Our designed multi-head attention aggregator achieves the best results, whereas the mean aggregator performs the worst. This highlights the effectiveness of attention mechanisms in discerning and weighting the importance of contextual information.

\autoref{fig:hyperparameter_analysis}(c) demonstrates the effect of varying hop counts on the graph. The results indicate that SARMP achieves competitive performance at the first hop and optimal performance at the second hop. Further increasing the hop count, however, leads to performance degradation. This phenomenon is likely attributable to the exponential increase in neighbor sampling with deeper hops, which consequently introduces more noise and potentially less relevant information.

\subsection{Parametric Analysis: RQ3}

\autoref{tab:parameters_fb15k237} details the parameter counts for each model on the FB15k-237 dataset.
SARMP exhibits significantly greater storage efficiency than embedding-based methods, as it obviates the need to calculate and store entity embeddings.
Its model complexity is $\mathcal{O}(n \cdot K^{hops})$.
Furthermore, SARMP proves more effective than GNN-based approaches like PathCon.
SARMP can achieve superior performance with fewer hops and a smaller sampled neighborhood, and notably, it does not require the explicit modeling of complex paths from head to tail entities, unlike PathCon.

\begin{table}
  \centering
  \caption{Number of parameters of all models on FB15k-237.}
  \label{tab:parameters_fb15k237}
  {
    \setlength{\tabcolsep}{5pt}
  \begin{tabular}{c|cccccc} 
    \toprule
    \textbf{Model}   & \textbf{TransE} & \textbf{DisMult} &\textbf{RotatE}  & \textbf{QuatE} & \textbf{PathCon} & \textbf{SARMP}  \\
    \midrule
    Param & 5.9M   & 5.9M    & 11.7M   & 23.6M & 1.67M   & \textbf{0.34M }\\ 
    \bottomrule
  \end{tabular}
  } 
\end{table}

\section{Conclusion}
In this paper, we introduced a novel semantic-aware relational message passing framework for Knowledge Graph Completion, designed to overcome noise and information dilution in traditional methods.
Our approach features a semantic-aware Top-K neighbor selection strategy to identify the most relevant incident edges and a specialized multi-head attention aggregator to effectively fuse their information, generating context-rich node representations.
This improved alternating message passing scheme more accurately captures and propagates crucial semantic cues.
Comprehensive experiments on benchmark datasets demonstrate that our method significantly outperforms existing state-of-the-art approaches, confirming its effectiveness and robustness.

\section{Conflict of Interest Statement}
The authors have no competing interests.

%
%

%
%
%
%

\end{document}